\documentclass[journal,twocolumn]{article}
\pdfoutput=1
\usepackage{cite,url}
\usepackage{amsmath,amssymb,amsfonts}
\usepackage{algorithmic}
\usepackage{graphicx}
\usepackage{textcomp}
\newcommand{\diff}{\mathrm{d}}
\newcommand{\minimize}{\mathop{\rm minimize}\limits}
\newcommand{\maximize}{\mathop{\rm maximize}\limits}

\begin{document}
\title{Aortic root landmark localization with optimal transport loss for heatmap regression
\thanks{This work has been submitted to the IEEE for possible publication. Copyright may be transferred without notice, after which this version may no longer be accessible.}}
\author{Tsuyoshi Ishizone
\thanks{Corresponding authors: T. Ishizone and M. Miyasaka}
\thanks{T. Ishizone is with the Meiji Institute for Advanced Study of Mathematical Sciences, Meiji University, 4-21-1 Nakano, Nakano-ku, Tokyo, Japan (e-mail: tsuyoshi.ishizone@gmail.com).}, 
Masaki Miyasaka
\footnotemark[2]
\thanks{M. Miyasaka and S. Ochi are with the Department of Laboratory Medicine, The Jikei University School of Medicine, 3-25-8 Nishi-shinbashi, Minato-ku, Tokyo, Japan (e-mail: masaki108@gmail.com, ochisae1024@jikei.ac.jp).}
\thanks{M. Miyasaka and N. Tada are with the Cardiovascular Center, Sendai Kousei Hospital, 1-20 Tsutsumidori-Amamiyamachi, Aoba-ku, Sendai-shi, Miyagi, Japan. (e-mail: masaki108@gmail.com, noriotada@gmail.com).}, 
Sae Ochi
\footnotemark[4], 
Norio Tada
\footnotemark[5], and 
Kazuyuki Nakamura
\thanks{K. Nakamura is with the School of Interdisciplinary Mathematical Sciences, Meiji University, 4-21-1 Nakano, Nakano-ku, Tokyo, Japan (e-mail: knaka@meiji.ac.jp).}.
}
\date{}

\maketitle

\begin{abstract}
Anatomical landmark localization is gaining attention to ease the burden on physicians.
Focusing on aortic root landmark localization, the three hinge points of the aortic valve can reduce the burden by automatically determining the valve size required for transcatheter aortic valve implantation surgery.
Existing methods for landmark prediction of the aortic root mainly use time-consuming two-step estimation methods.
We propose a highly accurate one-step landmark localization method from even coarse images.
The proposed method uses an optimal transport loss to break the trade-off between prediction precision and learning stability in conventional heatmap regression methods.
We apply the proposed method to the 3D CT image dataset collected at Sendai Kousei Hospital and show that it significantly improves the estimation error over existing methods and other loss functions.
Our code is available on GitHub. \footnote{https://github.com/ZoneTsuyoshi/grid-based-lipschitz-penalty}
\end{abstract}


\section{Introduction}
\label{sec:introduction}
Anatomical landmark detection is a pivotal task in medical image analysis and is essential for a multitude of clinical and research applications. 
Accurate landmark localization aids in identifying pathological changes, planning surgical interventions, and aligning multimodal images for comprehensive analysis. 

Landmark localization for the aortic root from 3-dimensional computed tomography (CT) is crucial due to its significant implications for cardiovascular diagnostics and interventions. 
Precise localization of aortic root landmarks is essential for accurate measurement of dimensions, such as the aortic annulus, sinuses of Valsalva, and sinotubular junction, which are critical for diagnosing aortic diseases like aneurysms, dissections, and valve pathologies. 
These landmarks are also vital for planning and guiding interventions such as transcatheter aortic valve implantation (TAVI), where precise measurements ensure optimal prosthesis selection and positioning, reducing the risk of complications \cite{Jilaihawi2012-miyasaka1,maeno2017optimal-miyasaka4}. 
Moreover, reliable landmark detection aids in longitudinal studies by providing consistent reference points for tracking disease progression or treatment outcomes, ultimately improving patient care and advancing cardiovascular research.

The existing methods of landmark localization for aortic root are based on a two-stage estimation with global and local predictions \cite{al18-coronial-walk,al20-reinforcement-learning,tan19-stacom,noothout20-tmi}.
Global prediction predicts the approximate location of landmarks from downsampled low-resolution image data.
Local prediction extracts the original high-resolution image around the global prediction location and predicts the fine-scale location of landmarks.
A two-stage estimation method is often used because 3D CT images have large image sizes, and commonly used GPU memory cannot handle the computation.
However, building a model for a new dataset using these methods is expensive because two-stage estimation is often expensive for model training and hyperparameter search.

To solve this problem, this paper proposes a highly accurate one-step estimation method.
The proposed method consists of heatmap regression using U-Net \cite{ronneberger15_u-net} and optimal transport loss \cite{PeyreCuturi2019_ot}.
Heatmap regression is a commonly used framework for anatomical landmark prediction and is formulated as a problem of predicting a Gaussian heatmap centered on the correct landmark location from an input image.
Heatmap regression is known to have a trade-off between learning stability and prediction precision, depending on the value of the standard deviation of the Gaussian heatmap.
Although the precision increases when the standard deviation is low, learning tends to be unstable.
Learning instability is caused by class imbalance due to a larger proportion of voxels with no landmarks.
A high standard deviation makes learning more stable but decreases the prediction precision.
Although class imbalance is less likely to occur, the prediction of landmark existence will be wider, and the exact landmark locations will be ambiguous.
Existing two-stage heatmap regression methods use a larger standard deviation in the first stage to get rough landmark locations and a smaller standard deviation in the second stage to increase the prediction precision.
The proposed method resolves the trade-off between learning stability and prediction precision by using optimal transport loss for heatmap regression \cite{li23-ssa-standard-dev,luo21-rethinking-heatmap}.
The binary classification problem, derived as the dual problem of the optimal transport problem, is less sensitive to class imbalance by adjusting the weights of samples from each class \cite{serrurier21-ot-classification,villani09-optimal}.
Our proposed GLiP is an optimal transport loss function designed for heatmap regression.


In summary, our contributions are as follows:
\begin{itemize}
    \item We propose a new loss function suitable for heatmap regression and predict the heatmap using U-Net.
    Just as the Wasserstein GAN stabilizes the learning of GANs, the proposed loss stabilizes the learning of heatmap regression by U-Net.
    The loss breaks the trade-off between learning stability and prediction precision and can achieve stable and accurate predictions.
    \item We apply the proposed method to a 3D CT image dataset collected at Sendai Kousei Hospital and compare and evaluate it with existing methods.
    We show that the proposed method localizes landmarks with such high accuracy that the deviation from a coarse image with a resolution of 1.6 mm can be suppressed to about one voxel.
\end{itemize}

The rest of this paper is organized as follows.
We review previous studies in Section~\ref{sec:related-works} and describe the details of the proposed method in Section~\ref{sec:method}.
Section~\ref{sec:experiment} describes our experiments using 3D CT images, and Section~\ref{sec:conclusion} summarizes the paper.

\section{Related Works}
\label{sec:related-works}
\subsection{Anatomical Landmark Detection}
Anatomical landmark prediction has been developed for various medical imaging data, such as CT \cite{lang22_ld_ct_craniomaxillofacial,palazzo20_ct_cranio,ma20-shift-equivalent-ct,ghesu19-reinforcement-ct,liu23-bone-ct-densenet,chen22_ld_ct_cephalometric_sa-lstm,qian20_ct_cepha}, magnetic resonance image \cite{jain20_mr_head,kundeti20_ld_rl_mri_brain,tan23-adversarial-cerebrovascular,kundeti20_ld_mri_brain},  X-ray \cite{zhou21_ld_xray_cephalometric,gai23_ld_xray_limb,du22-xray-cepha-attention,liu20-pelvis-xray,lu23-pelvis-xray,dai19_ld_xray_cephalo,xiao23_ld_xray_knee,zhou21-ultrasound-xray-reinforcement-learning,zhang23-xray-hip-iiesc-net,reddy21-xray-cepha-spine-attention,millan-arias23-xray,bai21-xray-bone,lu22-xray-cepha-gcn,tiulpin19_radio_knee,kwon21-cepha-radio}, ultrasound \cite{zhou21-ultrasound-xray-reinforcement-learning,chen20-ultrasound-graph}, echocardiography \cite{wang21-echo-rnn,tian21-echo-period,wan23_ld_ventricle_echo,andreassen22-annulus-segmentation}, laparoscopes \cite{pozdeev21_ld_laparoscopic}, esophagogastroduodenoscopy \cite{lopes22-gastric-egd}, intraoral scans \cite{wu22-intraoral-scan-tooth,wang24-teeth-point-cloud}.

Landmark localization methods are divided into three streams: classification-based, exploring-based, and regression-based methods.
Some methods use bounding-box estimation \cite{lang22_ld_ct_craniomaxillofacial,TAHOCES21-bouding-box,chen20-ultrasound-graph} and data augmentation \cite{kundeti20_ld_mri_brain,lopes22-gastric-egd,frueh22-realtime-cross-subjects} before localizing landmarks.
Classification-based methods classify whether a landmark exists at particular voxels \cite{Oktay2017StratifiedDF,Zheng2012AutomaticAortaSegmentation,tan22-cerebrovascular-mra-segmentation,chen20-ultrasound-graph,Astudillo20-segmentation,tan22-cerebrovascular-mra-segmentation,tan21-regression-tree-bifurcation,gai23_ld_xray_limb,liu23-bone-ct-densenet,lu21-contour-transformer-segmentation}.
Exploring-based methods gradually explore landmark positions with walking particles and reinforcement learning \cite{al18-coronial-walk,al20-reinforcement-learning,Jung2015ForestWM,zhou21-ultrasound-xray-reinforcement-learning,jain20_mr_head,kundeti20_ld_rl_mri_brain,ghesu19-reinforcement-ct}.
Typical reinforcement learning methods use states - the patches of the images, actions - position move, and rewards - the distances between the current position and the ground truth landmark \cite{al20-reinforcement-learning,jain20_mr_head,kundeti20_ld_rl_mri_brain,ghesu19-reinforcement-ct}.
These methods use deep Q-network (DQN) \cite{Mnih2015HumanLevelControl-deepq,ghesu19-reinforcement-ct,jain20_mr_head,kundeti20_ld_rl_mri_brain}, double DQN \cite{Hasselt2016DoubleQLearning,kundeti20_ld_rl_mri_brain}, and dueling DQN \cite{Wang2015DuelingNetwork,kundeti20_ld_rl_mri_brain} as arachitectures.
As an atypical method, Zhou {\it et al.} \cite{zhou21-ultrasound-xray-reinforcement-learning} consider the heatmap generating method as a state, explore the generating method, and learn the heatmap regression task as a reward.

Regression-based methods are the mainstream for localizing landmarks.
The stream has two sub-streams: vector regression methods and heatmap regression methods.
Vector regression methods predict landmark coordinates or offset from current grid to landmark positions \cite{Oktay2017StratifiedDF,Gall2011HoughForests,Sun2012ConditionalRegressionForests,sanchez24-xray-segmentation-regression-limb,tan21-regression-tree-bifurcation,chen22_ld_ct_cephalometric_sa-lstm,gai23_ld_xray_limb,tiulpin19_radio_knee,lu23-pelvis-xray,du22-xray-cepha-attention,ma20-shift-equivalent-ct,liu23-bone-ct-densenet,kwon21-cepha-radio,lu22-xray-cepha-gcn}. 
Ma {\it et al.} \cite{ma20-shift-equivalent-ct} uses the shift-equivalent property of the localizer network for end-to-end training of high-resolution images.
Sanchez {\it et al.} \cite{sanchez24-xray-segmentation-regression-limb} presents segmentation-guided coordinate regression, which consists of U-Net and VGG architectures.
Lu {\it et al.} \cite{lu22-xray-cepha-gcn} uses graph convolutional networks to extract relationships among the coordinates of the landmarks.

Heatmap regression methods construct heatmap centered on ground truth landmarks and predict the heatmap \cite{tan19-stacom,Payer2016RegressingHF,tan22-cerebrovascular-mra-segmentation,tan21-regression-tree-bifurcation,chen22_ld_ct_cephalometric_sa-lstm,qian20_ct_cepha,wan23_ld_ventricle_echo,zhou21-ultrasound-xray-reinforcement-learning,reddy21-xray-cepha-spine-attention,dai19_ld_xray_cephalo,xiao23_ld_xray_knee,zhang23-xray-hip-iiesc-net,bai21-xray-bone,tan23-adversarial-cerebrovascular,fu22_ld_image_face,kundeti20_ld_mri_brain,wang21-echo-rnn,tian21-echo-period,liu23-bone-ct-densenet,guo23-4d-motion,millan-arias23-xray,luo21-fovea,andreassen22-annulus-segmentation}.
Tan {\it et al.} \cite{tan21-regression-tree-bifurcation} predicts bifurcation landmarks with heatmaps, segmentation, and orientation regression.
Chen {\it et al.} \cite{chen22_ld_ct_cephalometric_sa-lstm} simultaneously learns coarse heatmap regression and fine coordinate regression.
Tan {\it et al.} \cite{tan22-cerebrovascular-mra-segmentation} combines heatmap regression with semantic segmentation and classification for localizing cerebrovascular landmarks.
Qian {\it et al.} \cite{qian20_ct_cepha} and Wan {\it et al.} \cite{wan23_ld_ventricle_echo} propose new loss functions to balance the foreground and background pixels.
Xiao {\it et al.} \cite{xiao23_ld_xray_knee} uses graph convolutional networks to feature relationships among landmarks.
Fu {\it et al.} \cite{fu22_ld_image_face} uses transfer learning for facial landmark detection to apply to a small dataset of fetal alcohol spectrum disorders.
Millan-Arias {\it et al.} \cite{millan-arias23-xray} utilizes X-ray images obtained from different machines.
Tan {\it et al.} \cite{tan23-adversarial-cerebrovascular} uses cross-modality information of MRA and CTA for adversarial training of heatmaps.
Dai {\it et al.} \cite{dai19_ld_xray_cephalo} uses distance maps instead of heatmaps and uses image gradient difference loss \cite{nie18-image-gradient-difference-medical-image-synthesis} for maintaining the sharp edge.
Our proposals also belong to heatmap regression methods.

Some methods focus on localizing landmarks of the aortic root from 3D-CT dataset \cite{tan19-stacom,noothout20-tmi,Astudillo20-segmentation,al18-coronial-walk,al20-reinforcement-learning,TAHOCES21-bouding-box}.
Tan {\it et al.} \cite{tan19-stacom} combines global heatmap regression with U-Net and local directional vector regression with CNN.
Noothout {\it et al.} \cite{noothout20-tmi} use directional vector regression and landmark affiliation prediction with ResNet in global and local stages.
Astudillo {\it et al.} \cite{Astudillo20-segmentation} estimates landmarks via semantic segmentation using DenseVNet.
Al {\it et al.} \cite{al18-coronial-walk,al20-reinforcement-learning} uses the coronial walk and deep reinforcement learnings to explore the landmark position.
Tahoces {\it et al.} \cite{TAHOCES21-bouding-box} estimates landmarks via bounding box estimation.

These methods consist of multiple stages of learning and require tuning at each stage.
Typically, they are divided into two stages: global and local, where global estimation is used to identify areas where landmarks are located, and local estimation is used to output final estimates of landmarks.
To compensate for the lack of GPU memory, these methods reduce the image size by downsampling for global estimation and use the original resolution image by extracting patches for local estimation.
However, multi-stage network configurations require exploring many elements, such as hyperparameters and modules, at each stage, which increases the exploring cost.
To avoid these, our proposed method accurately predicts landmarks using only downsampled low-resolution images.
The proposed method is also suitable for situations where only low-resolution CT images are available due to scanner or patient problems.

\subsection{Optimal Transport}
\label{ssec:rw-optimal-transport}
Optimal transport is a distance measure between probability distributions, formulated as the cost of `optimal transport' between them.
Optimal transport type losses have become a widely used alternative to Kullback-Leibler type losses because of the stability of learning \cite{Arjovsky17-WassersteinGAN,gulrajani17-nips-wgan-gp,roy24-drlgan,zhu2023uniform-wasserstein-sgd}.
A typical example is the Wasserstein GAN \cite{Arjovsky17-WassersteinGAN}, a development of the GAN \cite{Goodfellow2014-GAN,goodfellow20-gans-acm}.
GAN is a deep generative model that minimizes the Jensen-Shannon (JS) divergence similar to KL. 
WGAN stabilizes the learning of GAN by replacing the JS divergence with the optimal transport Wasserstein distance.
The optimal transport that attracted attention in WGAN is used in various fields, such as time-series analysis \cite{cazelles21-wasserstein-fourier-distance}, variable screening \cite{jeong2023wasserstein-variable-screening}, out-of-distribution detection \cite{wang24-wasserstein-wood-ood}, and object detection \cite{Han_2020_wasserstein-object-detection}, as an alternative to cross-entropy loss based on KL.
We propose a loss function based on optimal transport to solve the class imbalance problem in conventional heatmap regressions.

\begin{figure*}
    \centering
    \includegraphics[width=0.7\textwidth]{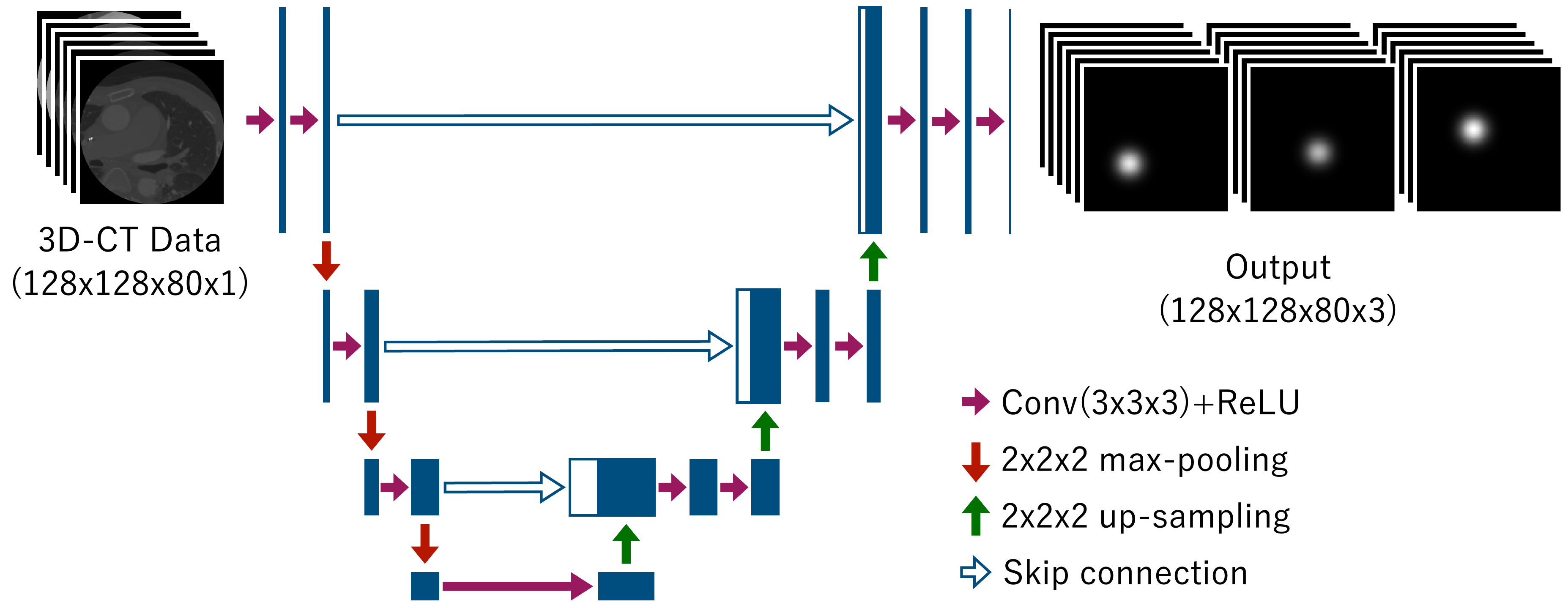}
    \caption{Architecture of U-Net. The input is 3D-CT data, and the output is the predicted heatmaps of three landmarks. {\it Magenta arrows} denote convolution operation with kernel size 3 and ReLU activation function. {\it Red arrows} denote the max-pooling of size 2, {\it green arrows} denote the up-sampling of size 2, and {\it blue blanked arrows} denote the skip connection. {\it Blue boxes} correspond to the size of features. The width and height of the boxes are proportional to feature dimension and image size.}
    \label{fig:u-net}
\end{figure*}

\section{Method}
\label{sec:method}
Our architecture is a heatmap regression based on U-Net, as shown in Figure~\ref{fig:u-net}.
The method inputs a 3D CT image and predicts landmark locations as a heatmap.
First, an overview of the architecture is given, followed by a description of how the heatmaps are created.
The loss function used in the heatmap regression, which is the heart of the method, is based on the optimal transport distance.
After an overview of the optimal transport distance, we propose a loss that modifies the Lipschitz penalty constraints for heatmap regression.

\subsection{U-net}
\label{ssec:u-net}
The basis of our architecture is U-net, a well-known convolutional neural network mainly used in image segmentation tasks \cite{ronneberger15_u-net,zhou2018unet++,yang2019drunet,zunair2021sharp-unet,ibtehaz2020multiresunet,deng22-elu-net,lu2022half-unet,punn2021modality-unet-survey}.
The architecture is structured as an encoder-decoder network with a symmetric layout.
The convolution layers in the encoder extract local information from the input.
The max-pooling layers aggregate local information and decrease voxel size.
The upsampling layers expand the features from global to local.
The skip connections provide a pathway for gradients during training, which helps combat the vanishing gradient problem \cite{he2016resnet,li18-resnet-adjustable,he20-resnet-generalize}.
This sophisticated architecture provides robust estimation even for noisy or obscured images.

\subsection{Heatmap Generation}
\label{ssec:heatmap}
Ground truth heatmaps are created from landmarks manually annotated by experts.
The heatmap for each landmark is the same size as the input image and has the highest value at the landmark position.
We use the standard Gaussian function and give the heatmap value of the $i$-th landmark of $b$-th sample at voxel $v$ as
\begin{equation}
    h_{v}^{b,i}=\exp\left\{-\frac{\|\boldsymbol{g}^b_v-\boldsymbol{l}^{b,i}\|}{2\sigma^2}\right\},
\end{equation}
where $\sigma$ is the standard deviation, $\boldsymbol{g}^b_v\in\mathbb{R}^3$ represents the position of the voxel $v$ of the $b$-th sample, $\boldsymbol{l}^{b,i}\in\mathbb{R}^3$ represents the position of the $i$-th landmark of the $b$-th sample.

\subsection{Optimal Transport}
\label{ssec:optimal-transport}
Optimal transport (OT) is a mathematical framework that has been significantly applied in machine learning, particularly in distribution comparison and alignment areas \cite{Villani2008_ot,PeyreCuturi2019_ot}. 
The method provides a way to quantify the distance between two probability distributions meaningfully and geometrically intuitively.
The theory seeks to find the most cost-effective way of transforming one distribution into another.
In its most common form, the optimal transport problem can be expressed as
\begin{align}
    &\minimize_{\pi\in\Pi(\mu,\nu)}\int_{\mathcal{X}\times\mathcal{Y}}C(\boldsymbol{x},\boldsymbol{y})\diff\pi(\boldsymbol{x},\boldsymbol{y}),\\
    &\Pi(\mu,\nu)=\left\{\pi:\mathcal{X}\times\mathcal{Y}\to[0,1]\right|\notag\\
    &\qquad\qquad\int_\mathcal{Y}\pi(\boldsymbol{x},\boldsymbol{y})\diff \boldsymbol{y}=\mu(\boldsymbol{x}),\notag\\
    &\qquad\qquad\left.\int_\mathcal{X}\pi(\boldsymbol{x},\boldsymbol{y})\diff \boldsymbol{x}=\nu(\boldsymbol{y})\right\},
\end{align}
where $\mathcal{X}$ and $\mathcal{Y}$ are topological spaces, $\pi:\mathcal{X}\times \mathcal{Y}\to[0,1]$ represents a transport plan (coupling), which is a joint distribution whose marginals are $\mu:\mathcal{X}\to[0,1]$ (source distribution) and $\nu:\mathcal{Y}\to[0,1]$ (target distribution).
The $C:\mathcal{X}\times \mathcal{Y}\to\mathbb{R}$ is a cost function that measures the cost of transporting mass from $\boldsymbol{x}\in\mathcal{X}$ to $\boldsymbol{y}\in\mathcal{Y}$.
$\Pi(\mu,\nu\}$ is the set of all possible transport plans (couplings) that have $\mu$ and $\nu$ as their marginals.

Assuming that the space $\mathcal{X}=\mathcal{Y}$ in which the probability measure is defined and that the cost function $C$ is represented by a distance function $d:\mathcal{X}\times\mathcal{X}\to\mathbb{R}$, the optimal transport problem can be expressed as
\begin{equation}
    \minimize_{\pi\in\Pi(\mu,\nu)}\int_{\mathcal{X}\times\mathcal{X}}d(\boldsymbol{x},\boldsymbol{y})\diff\pi(\boldsymbol{x},\boldsymbol{y}).
\end{equation}
The dual problem of this problem is expressed by
\begin{equation}
    \maximize_{\phi\in\mathcal{C}_b^L(\mathcal{X})}\int_\mathcal{X}\phi(\boldsymbol{x})\diff\mu(\boldsymbol{x})-\int_\mathcal{X}\phi(\boldsymbol{y})\diff\nu(\boldsymbol{y}),\label{eq:ot-dual}
\end{equation}
where $\mathcal{C}_b^L(\mathcal{X})=\{\phi\in\mathcal{C}(\mathcal{X})|\phi:\text{bounded, 1-Lipschitz}\}$ represent the set of bounded and 1-Lipschitz continuous functions from $\mathcal{X}$ to $\mathbb{R}$.

This problem is solved by increasing the output value of $\phi(\boldsymbol{x}^\mu)$ in the sample $\boldsymbol{x}^{\mu}$ from distribution $\mu$ and decreasing the value of $\phi(\boldsymbol{x}^\nu)$ in the sample $\boldsymbol{x}^\nu$ from distribution $\nu$, which increases the interobjective value.
Considering these two distributions as positive and negative example distributions, the problem can be considered a binary classification problem.

When $\mathcal{X}$ is a continuous space, the binary classification problem with optimal transport is formulated as a problem of finding the optimal parameter $\boldsymbol{\theta}$ that minimizes the objective function using a parametric function $\phi_{\boldsymbol{\theta}}$.
The gradient penalty method is used to satisfy the 1-Lipschitz constraint on the function $\phi$.
The method uses the fact that the norm of the gradient is 1 almost everywhere when using the optimal coupling $\pi^*$ \cite{gulrajani17-nips-wgan-gp}.
Sampling $\boldsymbol{x}$, $\boldsymbol{y}$ from each distribution $\mu$ and $\nu$ respectively, the penalty $(\|\nabla_{\boldsymbol{x}}\phi_{\boldsymbol{\theta}}(\tilde{\boldsymbol{x}})\|-1)^2$ is applied to the loss function such that the norm of the gradient of its interior point $\tilde{\boldsymbol{x}}=t\boldsymbol{x}+(1-t)\boldsymbol{y}$ is 1.

\subsection{Grid-based Lipschtiz Penalty (GLiP)}
\label{ssec:glip}

There are two challenges in applying the gradient penalty method to heatmap regression.
First, the method mainly applies to tasks classified by image and is unsuitable for tasks classified by voxel, such as heatmap regression.
Second, the method assumes that labels are binary and does not cover soft labels that take values between 0 and 1.
To address these issues, we propose GLiP (grid-based Lipschitz penalty), which imposes a Lipschitz constraint penalty for heatmap regression.
The penalty term of GLiP is formulated as
\begin{equation}
    \mathcal{L}^{\mathrm{Penalty}}(\boldsymbol{I})=\sum_{u,v\in E}\sum_{i=1}^{N_l}\left(|f_{\boldsymbol{\theta}}(\boldsymbol{I})^i_{u}-f_{\boldsymbol{\theta}}(\boldsymbol{I})^i_v|-\boldsymbol{1}\right)^2,
\end{equation}
where $I\in\mathcal{I}$ is the input image, $\mathcal{I}=\mathbb{R}^{H\times W\times D}$ is the image space, $E$ is the set of edges whose two grids are adjacent, $f_{\boldsymbol{\theta}}:\mathcal{I}\to\mathcal{I}^{N_l}$ is the NN based on U-Net, $f_{\boldsymbol{\theta}}(\boldsymbol{I})_v^i$ is the $i$-th output value at $v$-th grid, and $N_l$ denotes the number of landmarks.

Combined with the original optimal transport problem, the entire loss function of GLiP is expressed as
\begin{align}
    \mathcal{L}^{\mathrm{GLiP}}(\mathcal{B})&=\mathcal{L}^{\mathrm{OT}}(\mathcal{B})+\lambda\frac{1}{|\mathcal{B}|}\sum_{b\in\mathcal{B}}\mathcal{L}^{\mathrm{Penalty}}(\boldsymbol{I}^b),\label{eq:loss}\\
    \mathcal{L}^{\mathrm{OT}}(\mathcal{B})&=\frac{1}{|\mathcal{B}|N_l}\sum_{i=1}^{N_l}\left\{-\frac{\sum_{b\in\mathcal{B}} \boldsymbol{h}^{b,i}\cdot f_{\boldsymbol{\theta}}(\boldsymbol{I}^b)^i}{\sum_{b\in\mathcal{B}}\boldsymbol{h}^{b,i}\cdot\boldsymbol{1}}\right.\notag\\
    &\qquad\qquad\left.+\frac{\sum_{b\in\mathcal{B}} (\boldsymbol{1}-\boldsymbol{h}^{b,i})\cdot f_{\boldsymbol{\theta}}(\boldsymbol{I}^b)^i}{\sum_{b\in\mathcal{B}}(\boldsymbol{1}-\boldsymbol{h}^{b,i})\cdot\boldsymbol{1}}\right\}\label{eq:ot-loss},
\end{align}
where $\mathcal{B}$ is the minibatch set, $\boldsymbol{I}^b\in\mathcal{I}$ is the CT image of the $b$-th sample, $\boldsymbol{h}^{b,i}\in\mathcal{I}$ is the heatmap of the $i$-th landmark of the $b$-th sample, $f_{\boldsymbol{\theta}}(\boldsymbol{I}^b)^i\in\mathcal{I}$ is the $i$-th output values of the image $\boldsymbol{I}^b$.
The loss function of the optimal transport is also modified according to the value range [0,1] of the heatmap regression.

The relationship to the optimal transport problem described in Subsection~\ref{ssec:optimal-transport} is as follows.
In the binary classification problem \eqref{eq:ot-dual}, the distributions $\mu$ and $\nu$ correspond to the distributions of 1 and 0, respectively, where the landmarks are present or absent.
The values of $\mu$, $\nu$, and $\phi$ take values for voxel $v$ and landmark position $\boldsymbol{l}$ since the heatmap value corresponds to the probability of belonging to class 1 where the landmark is present.
Since the landmark position $\boldsymbol{l}$ depends on the CT image $\boldsymbol{I}$ and landmark type $i$, an element $\boldsymbol{x}$ of the topological space $\mathcal{X}$ can be expressed as the image $\boldsymbol{I}\in\mathcal{I}$, voxel $v\in\mathbb{N}^3$, and landmark type $i\in\mathbb{N}$: $\boldsymbol{x}=(\boldsymbol{I},v,i)\in\mathcal{X}$, $\mathcal{X}=\mathcal{I}\times\mathbb{N}^3\times\mathbb{N}$.
Since the image space $\mathcal{I}$ at training time can be reduced to the sample number space $\mathbb{N}$, $\mu$ and $\nu$ can be expressed as discrete measures $\mu(\boldsymbol{x})=h_v^{b,i}$ and $\nu(\boldsymbol{x})=1-h_v^{b,i}$.
The function $\phi$ to be learned can be expressed as $\phi_{\boldsymbol{\theta}}(\boldsymbol{x})=f_{\boldsymbol{\theta}}(\boldsymbol{I})_v^i$ using $f$ parameterized by $\boldsymbol{\theta}$.
From these arguments, the equation~\eqref{eq:ot-loss} is derived.

\section{Experiments}
\label{sec:experiment}

\subsection{Dataset}
\label{ssec:dataset}
Pre-TAVR CT DICOM data was available for 171 patients with severe aortic stenosis who were treated by TAVR at Sendai Kousei Hospital between January 2014 and June 2019. 
After excluding data for 4 patients with a bicuspid aortic valve, the remaining 167 datasets were used for further experiments.  
This study was approved by the local ethics committee (Approval No: 3-32).

CT measurement in this study was performed in accordance with the standard pre-TAVR CT measurement protocols \cite{Jilaihawi2012-miyasaka1,achenbach2013determination-miyasaka2}. 
Using the DICOM viewer Horos\textregistered version 3.3.6, three hinge points were identified, and their three-dimensional coordinates were recorded. 
For the analysis, experienced operators performed CT analysis to create the training data. \footnote{MM (M. Miyasaka) and SS (S. Suzuki) performed CT analysis to create the training data, with MM supervising SS. MM had experienced conducting research in the CT Core Lab at Cedars-Sinai Medical Center, which focused on structural heart disease and interventional cardiology imaging research \cite{Jilaihawi2012-miyasaka1,kawamori2018computed-miyasaka3,maeno2017optimal-miyasaka4,miyasaka2019clinical-miyasaka5}. The total volume of CT data we analyzed for pre-TAVR procedural planning exceeded 1,000 cases.}

Each CT data consisted of 320 images with 512$\times$512 slices to include the aortic valve.
Since the resolution of the slice images varied from 0.25 to 0.5 for each data, linear completion was performed to unify the spacing to 0.4.
Since the data size was too large to be computationally feasible on the GPU, 4-fold downsampling was performed in each direction in three dimensions to obtain 128$\times$128$\times$80 images.

We set the prediction targets for the three landmarks on the aortic valve, the hinge points of RCC, LCC, and NCC.
Heatmaps were created using the method described in Subsection \ref{ssec:heatmap} based on the positions, and training was performed as a heatmap regression task.

\subsection{Implemention Details}
\label{ssec:implemention}
The dataset was randomly divided 4:1 into CV and test datasets, and a 4-fold cross-validation was performed using the CV dataset.
Adam optimizer \cite{Kingma15-adam-iclr} was used with a learning rate of 0.001.
The standard deviation of the Gaussian heatmap is $\{0.5, 1, 2, 4, 8, 16, 32, 64\}$ and the Lipschitz penalty coefficient is $\{1, 10, 100\}$. 
The hyperparameters were chosen based on the best CV score for the median distance error of the landmark.
For the test dataset, landmarks were ensemble predicted by the network used in each fold.

\subsection{Comparison with Existing Methods and Loss Functions}
\label{ssec:comparison}
Experiments comparing the two existing methods \cite{tan19-stacom,noothout20-tmi} and the U-Net with six loss functions were conducted to verify the effectiveness of the proposed framework.
Tan {\it et al.} \cite{tan19-stacom} uses a framework based on a similar U-Net architecture but with two-stage heatmaps and landmark regression prediction structure.
Noothout {\it et al.} \cite{noothout20-tmi} combines landmark presence probability prediction with vector regression.
In addition to GLiP, we used weighted cross-entropy loss (WCE), focal loss (FL) \cite{Lin17-focal-ICCV}, mean-squared error (MSE), L1 error (L1), and smoothed L1 error (SL1) as loss functions.

\begin{figure}
    \centering
    \includegraphics[width=0.95\linewidth]{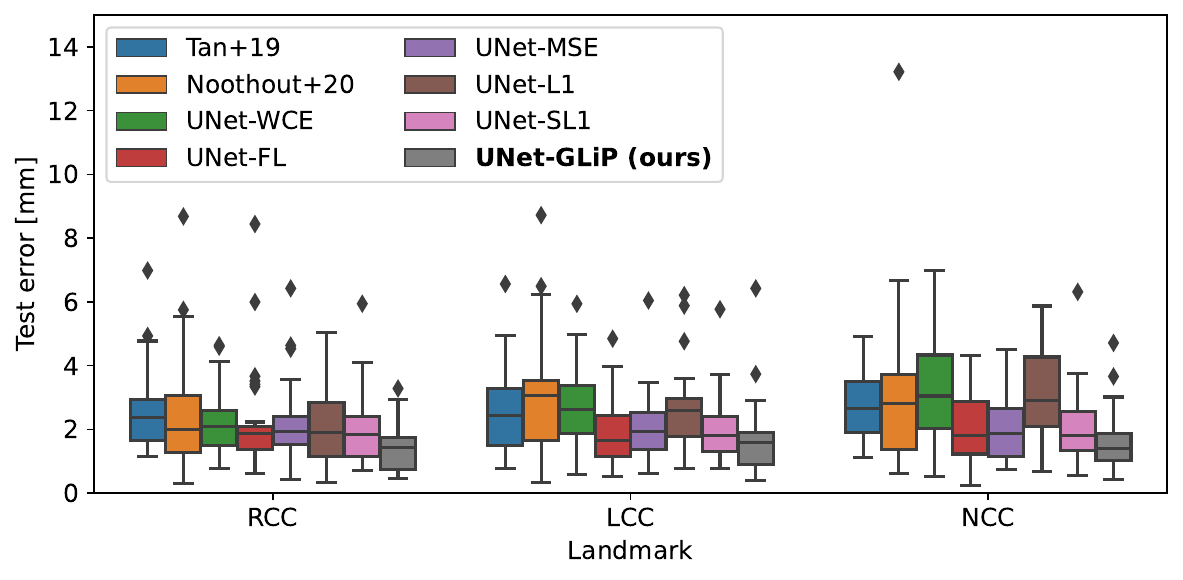}
    \caption{Boxplot of test Euclid distance error of the two existing methods (Tan+19 and Noothout+20) and six loss functions (WCE, FL, MSE, L1, SL1, GLiP) with U-Net. Although the worst errors of UNet-FL (47.7 mm) and UNet-SL1 (38.5 mm) are higher than 14 mm, these points are omitted for space limitation in the figure.}
    \label{fig:error}
\end{figure}

Figure~\ref{fig:error} shows a boxplot of the Euclid distance error for the existing method and heatmap regression with different loss functions for the test data set.
UNet-FL and UNet-SL1 have points outside the range of more than 14 mm, but these have been omitted due to space limitations in the figure.
The figure shows that the proposed method UNet-GLiP is the lowest at 0.25, 0.5, and 0.75 quantiles for all three hinge points.
UNet-FL, UNet-MSE, and UNet-SL1 are also the next lowest, indicating that the heatmap regression framework by U-Net is superior.

\begin{figure}
    \centering
    \includegraphics[width=0.95\linewidth]{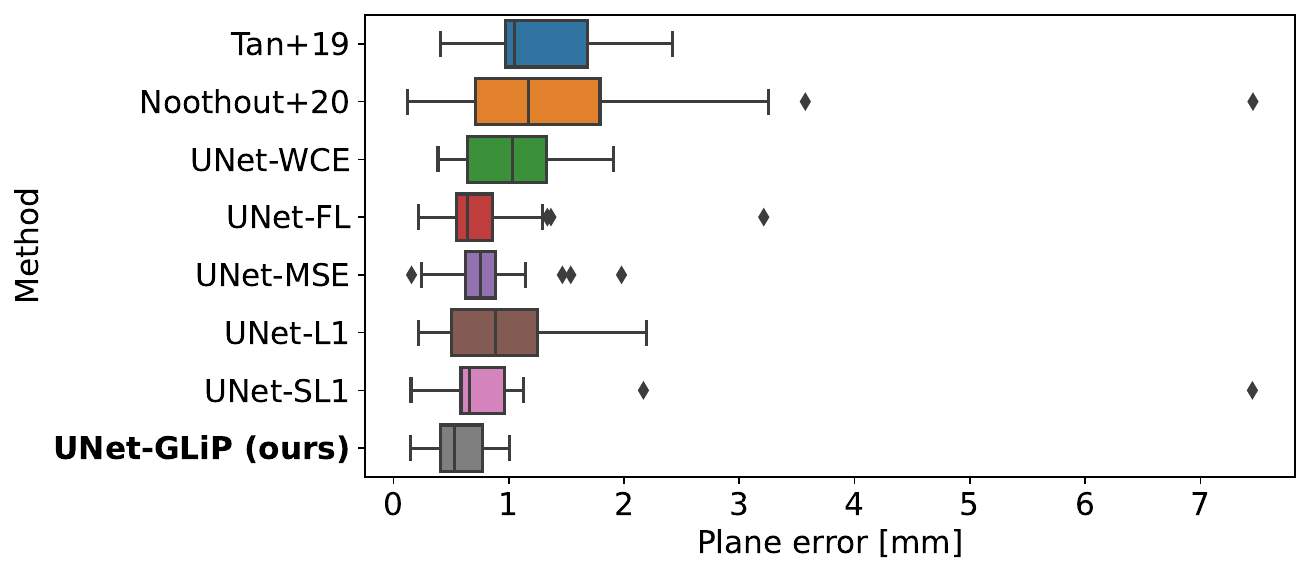}
    \caption{Boxplot of the test average projection distance error between the ground truth and predicted landmarks.}
    \label{fig:plane}
\end{figure}

\begin{figure}
    \centering
    \includegraphics[width=0.95\linewidth]{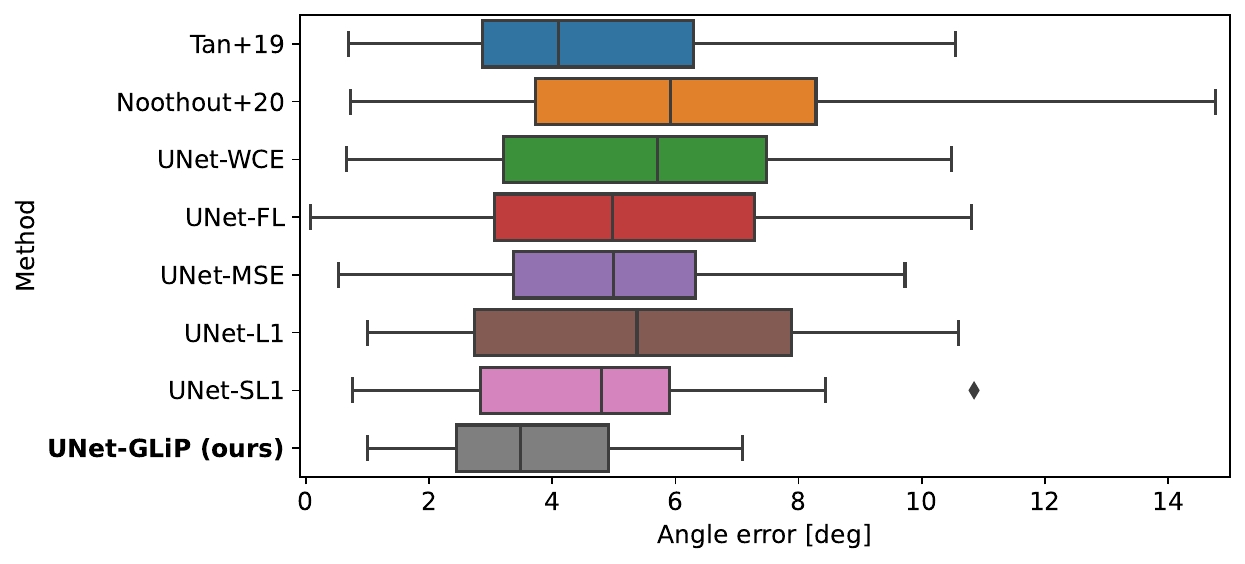}
    \caption{Boxplot of the test angle error between the ground truth and predicted planes. The worst error of Noohtout+20 and UNet-SL1 is higher than 15.}
    \label{fig:angle}
\end{figure}

We then evaluate the plane through which the three hinge points pass.
The plane contains the aortic annulus, the size of which is clinically important for determining the size of the prosthetic valve used in TAVR.
The aortic annulus size is determined by measuring the sizes of the aortic annulus on the virtual annulus plane. 
The virtual annulus plane is defined as the plane passing through the hinge points of the RCC, NCC, and LCC. 
The identification of hinge points and the measurement of the aortic annulus size are carried out in the following three steps \cite{achenbach2013determination-miyasaka2}.
\begin{enumerate}
    \item Rough adjustment: Using DICOM software, align the axis of the CT image of the aortic inlet perpendicular to the aortic valve. 
    Initially, rotate what was the axial plane and create an oblique plane that roughly approximates the orientation of the aortic valve.
    \item Fine adjustment: Move the plane up and down and rotate it to identify the three hinge points. 
    Adjust the plane to set it exactly through these three hinge points. 
    This plane is called the virtual annulus plane.
    \item Measurement: On the virtual annulus plane, draw a line around the aortic annulus to measure its size.
\end{enumerate}
If the predicted plane coincides with the correct plane, the CT analyst can skip steps 1 and 2 and draw a line around the aortic valve in the software.


To measure the degree of discrepancy between the correct and the estimated planes, we use two metrics: the average projection distance and the angle between the two planes.
The average projection distance is calculated as
\begin{align}
    &d^{PP}(\Pi(\boldsymbol{l}_1,\boldsymbol{l}_2,\boldsymbol{l}_3),\Pi(\hat{\boldsymbol{l}}_1,\hat{\boldsymbol{l}}_2,\hat{\boldsymbol{l}}_3))\notag\\
    &=\frac{1}{6}\sum_{i=1}^3\{d^{Pl}(\Pi(\hat{\boldsymbol{l}}_1,\hat{\boldsymbol{l}}_2,\hat{\boldsymbol{l}}_3),\boldsymbol{l}_i)+d^{Pl}(\Pi(\boldsymbol{l}_1,\boldsymbol{l}_2,\boldsymbol{l}_3),\hat{\boldsymbol{l}}_i)\},
\end{align}
where $\Pi:\mathbb{R}^3\times\mathbb{R}^3\times\mathbb{R}^3\to\mathcal{P}$ is the map from three points to the plane through the points, $\mathcal{P}$ is the set of planes in the three-dimensional space, $d^{Pl}:\mathcal{P}\times\mathbb{R}^3\to\mathbb{R}$ is the projection distance from a point to a plane.

A boxplot of the average projection distances and the angle between the ground truth and the predicted planes are shown in Figure~\ref{fig:plane} and Figure~\ref{fig:angle}, respectively.
The former figure shows the average projection distance of UNet-GLiP has the lowest worst-case error of approximately 1 mm, a level that can be practically handled.
The latter figure shows that UNet-GLiP has the smallest angular error in the 0.25, 0.5, and 0.75 quantiles.
The proposed method outperforms the existing methods regarding the two metrics.
This suggests that the proposed method is most effective for predicting the plane through which the three hinge points pass.



\subsection{Ablation Study}
\label{ssec:ablation}
Two experiments were conducted to verify the effectiveness of the proposed loss.

\subsubsection{Optimal Transport Loss without The Penalty Term}
\label{sssec:without-glip}
\begin{figure}
    \centering
    \includegraphics[width=0.95\linewidth]{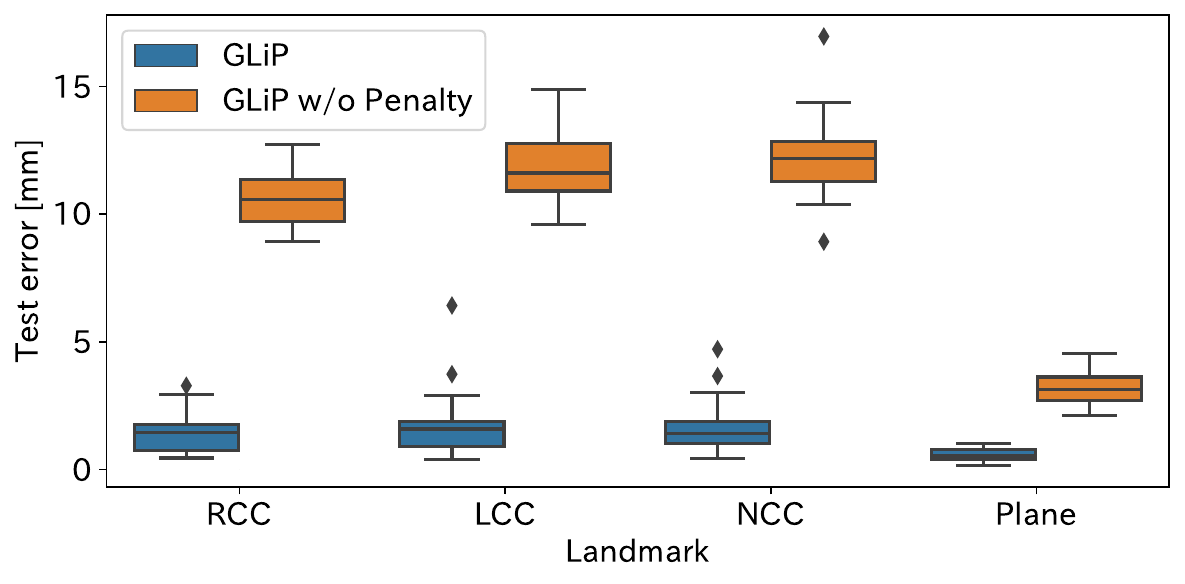}
    \caption{Boxplot of test Euclid distance error for optimal transport loss with/without the penalty term of GLiP. Plane means the average projection distance.}
    \label{fig:without-glip}
\end{figure}

To verify the validity of the Lipschitz constraint, experiments were conducted without the penalty term of GLiP.
A boxplot of the Euclid distance error for the test data set is shown in Figure~\ref{fig:without-glip}.
The figure shows that the error of GLiP without the penalty term becomes large.
This means that the constraint contributes to learning heatmap regression.

\subsubsection{Other Loss Functions with Lipschitz Constraint Penalty}
\label{ssec:with-gradient-penalty}
\begin{figure}
    \centering
    \includegraphics[width=0.9\linewidth]{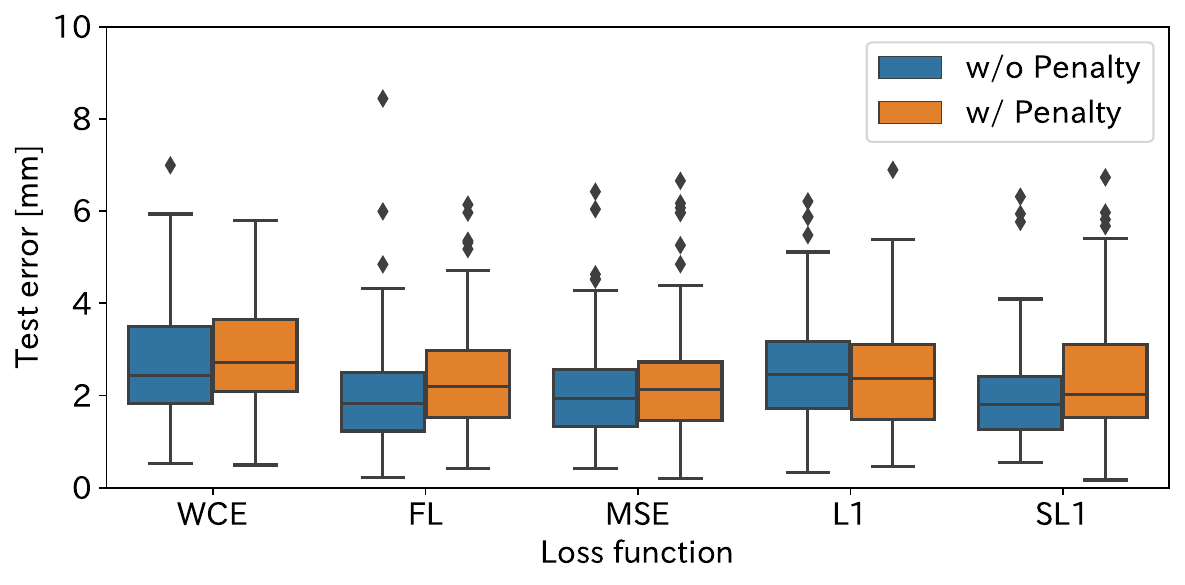}
    \caption{Boxplots of test Euclid distance error for each loss function with/without gradient penalty (GP). The worst errors for FL (47.7 mm) and SL1 (38.5 mm) without GP are omitted.}
    \label{fig:other-loss-with-gp}
\end{figure}

The Lipschitz constraint keeps the predicted difference between adjacent voxels constant, thus preventing the voxels from falling into flat predictions.
The constraint is derived from the optimal transport, but its constraint penalty may impact the proposed loss more than the optimal transport.
To confirm this effect, we performed an experiment in which a gradient penalty was added to loss functions other than the optimal transport loss.
Figure~\ref{fig:other-loss-with-gp} shows the boxplots of test error with and without gradient penalty (GP) for each loss function.
The worst errors for FL and SL1 without GP are omitted because they are large.
Except for the L1 loss, the values without GP are lower at 0.25, 0.5, and 0.75 quantiles.
This suggests that the Lipschitz constraint penalty improves prediction accuracy only for optimal transport losses.
These two ablation studies demonstrate that the combination of the optimal transport loss and the penalty constraint realized by GLiP provides highly accurate predictions.

\subsection{Extended Results}
\label{ssec:extended}

\subsubsection{Comparison for CT Quality}
\label{sssec:quality}
\begin{table}[tp]
    \centering
    \caption{Sample size by quality}
    \label{tab:quality}
    \begin{tabular}{l*{6}{r}}
        \hline
        Quality & 1 & 2  & 3- & 3+ & 4 \\
        \hline
        CV      & 3 & 33 & 64 & 31 & 2 \\
        Test    & 2 &  9 & 13 & 10 & 0 \\
        All     & 5 & 42 & 77 & 41 & 2 \\
        \hline
    \end{tabular}
\end{table}

\begin{figure}
    \centering
    \includegraphics[width=0.95\linewidth]{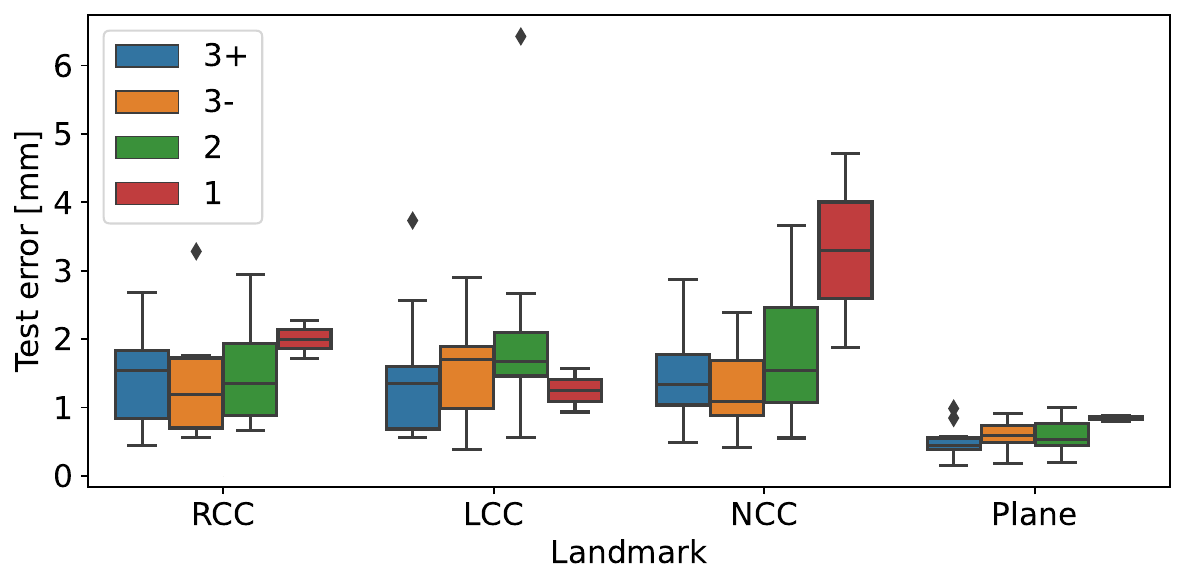}
    \caption{Boxplots of text Euclid distance error for each quality predicted by UNet-GLiP. Plane means the average projection distance.}
    \label{fig:quality}
\end{figure}

The quality of CT images is often influenced by technical, patient-related, or operational factors.
The data quality grading was performed semi-quantitatively and categorized into five grades. 
The quality was evaluated by each operator:
\begin{itemize}
    \item 1: Very Poor. Cannot be analyzed.
    \item 2: Poor. Difficult to analyze or measure accurately.
    \item 3-: Fair. Suitable for accurate analysis but slightly inferior to 3+.
    \item 3+: Good. Good quality for accurate measurement.
    \item 4: Excellent. Excellent quality for accurate measurement.
\end{itemize}

Figure~\ref{fig:quality} shows the error for each quality of UNet-GLiP on the test data.
It can be seen that the higher the data quality, the lower the error tends to be.
Surprisingly, the median error is below the expert-level error of 2 mm, even for the poor-quality data of grade 2.
Considering that the resolution of the CT images input to U-Net is 1.6 mm, the prediction is so accurate that it deviates by only one voxel level.

\subsubsection{Qualitative Results}
\label{sssec:qualitative}
\begin{figure}
    \centering
    \includegraphics[width=0.95\linewidth]{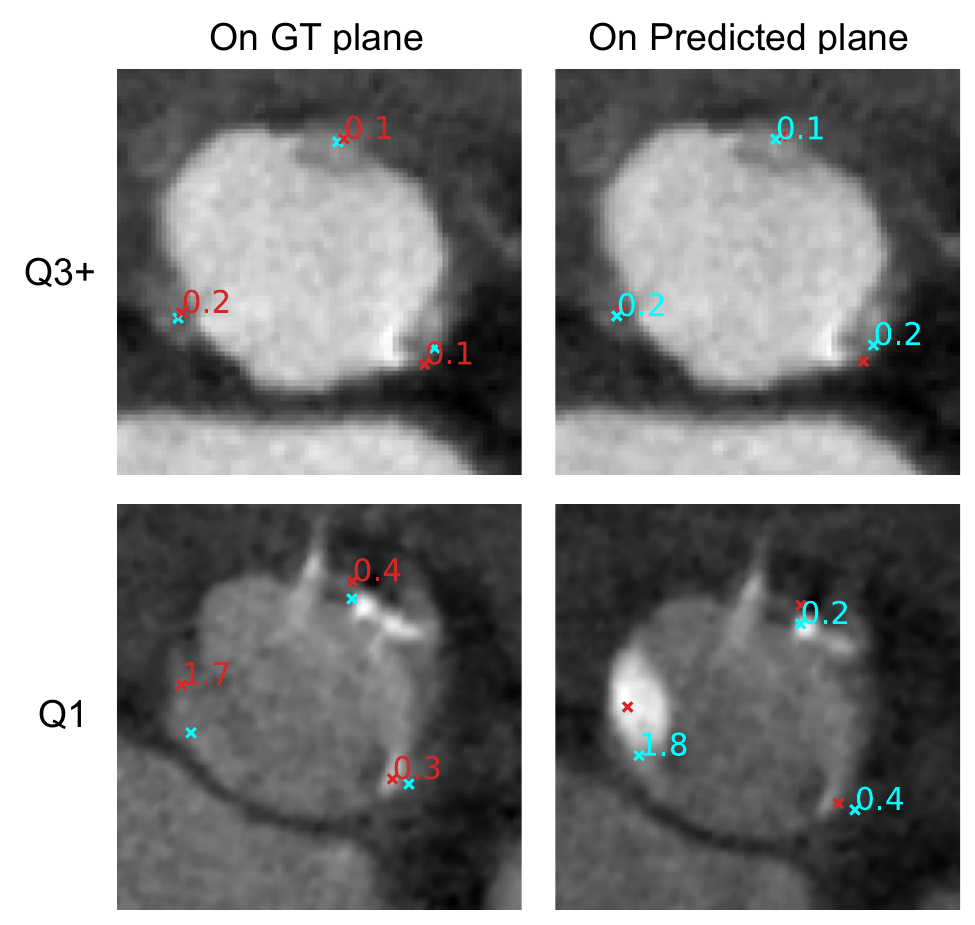}
    \caption{The ground truth landmarks ({\it light blue}) and the landmarks predicted by UNet-GLiP ({\it red}) on the ground truth plane (left columns) and the predicted plane (right columns) for data of quality 3+ (upper rows) and 1 (lower rows). In the left columns, predicted landmarks are projected onto the ground truth plane, and {\it red numbers} denote the projected distance (mm). In the right columns, ground truth landmarks are projected onto the predicted plane, and {\it light blue numbers} denote the projected distance (mm).}
    \label{fig:qualitative}
\end{figure}
Figure~\ref{fig:qualitative} shows the ground truth landmarks ({\it light blue}) and the landmarks predicted by UNet-GLiP ({\it red}).
The upper and lower rows show a test sample of quality 3+ and 1, respectively. 
The left column is plotted on the ground truth plane, and the right column is plotted on the prediction plane.
The ground truth plane is the plane through which the three ground truth landmarks pass. 
The predicted landmarks in the left column are projected onto the ground truth plane, and the numbers near them are the projected distances (mm).
Similarly, the ground truth landmarks are projected onto the prediction plane in the right column.
For the quality 3+ data in the top row, the deviations of the proposed method's predictions from the ground truth are so small that even experts cannot identify them.
For the quality 1 data in the bottom row, the lower left (NCC) prediction deviates from the ground truth.
Considering the input image's 1.6 mm resolution and the data's poor quality, this deviation is at a level that can occur even among experts.

\subsubsection{Second Stage Estimation}
\label{sssec:second-stage}
\begin{figure}
    \centering
    \includegraphics[width=0.9\linewidth]{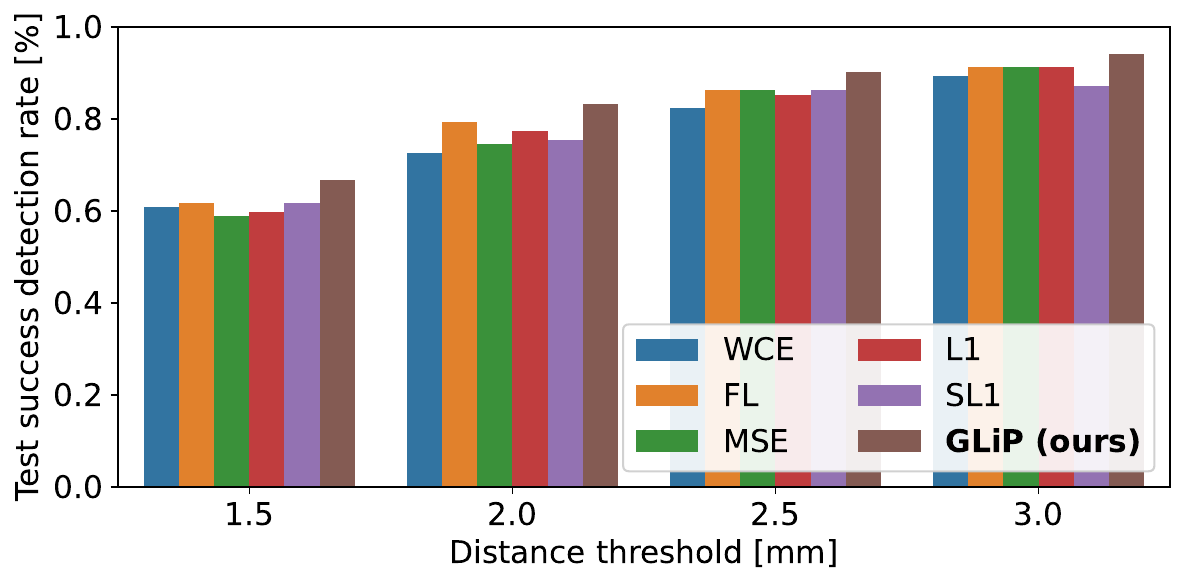}
    \caption{Success detection rate by the second-stage prediction with each loss function for test data}
    \label{fig:second-stage}
\end{figure}
\begin{figure}
    \centering
    \includegraphics[width=0.9\linewidth]{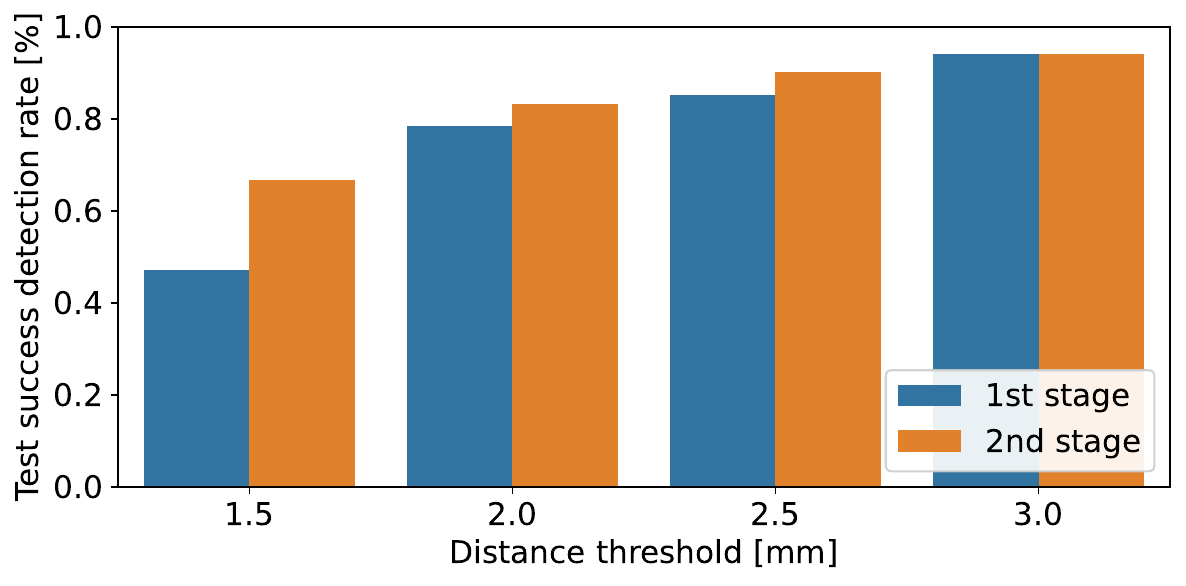}
    \caption{Success detection rate by the first/second prediction of UNet-GLiP for test data}
    \label{fig:second-stage-glip}
\end{figure}

Landmark prediction for medical images is mainly performed in two phases: global and local \cite{tan19-stacom,noothout20-tmi,Astudillo20-segmentation,TAHOCES21-bouding-box}.
This is because the original 3D CT image has a large number of pixels, so reducing the number of pixels is necessary to allow the GPU memory to accommodate the computation.
The global phase inputs the downsampled low-resolution image, while the local phase inputs a patch of high-resolution images around the global predictions.

To verify the effectiveness of the proposed method using the existing knowledge of two-stage prediction, we performed the local phase as the second-stage estimation.
For each loss function, the network with the smallest CV median error of landmarks was designated as the global network, and for each fold, the inputs to the local network were constructed from the outputs of the global network.
The input images were patched around the globally predicted landmarks with patch size $p^s=32$.
Disturbances were added to the patches to prevent overfitting where only the central voxel of global prediction is predicted in the local phase.
When the patch disturbance is $p^d$, a voxel region of $(p^s+2 p^d)^3$ around the global prediction landmark is extracted from the original CT image. 
A voxel region of $(p^s)^3$ starting at a uniform integer random number $\boldsymbol{u}\sim \mathrm{UI}[1,2p^d]^3$ is extracted from the $(p^s+2p^d)^3$ image.
Patch disturbances $p^d\in\{0,2,4,8\}$, heatmap standard deviation $\sigma\in\{0.5,1,2,4,8,16\}$, GLiP penalty $\lambda\in\{0.1, 1, 10\}$ were searched for minimizing the CV median distance error of landmarks.

Figure~\ref{fig:second-stage} shows the success detection rate in the second-stage landmark prediction for the test data for each loss function.
The success detection rate is the probability that the predicted landmark error is below the distance threshold.
The proposed loss has the highest success detection rate, suggesting that the second-stage prediction is stable and highly accurate.

Figure~\ref{fig:second-stage-glip} shows the success detection rate of landmarks in the first and the second-stage predictions by UNet-GLiP.
The success detection rate for the second-stage prediction is consistently higher, indicating the usefulness of the two-stage prediction.
The input image for the first-stage prediction has a resolution of 1.6 mm, so the largest discrepancy in the success detection rate from the second to the first is at the distance threshold of 1.5 mm.
The first-stage prediction is accurate enough for the expert level of 2 mm or less.
Two-stage prediction methods require appropriate hyperparameter adjustment at each stage, which increases the cost of hyperparameter adjustment and the cost of inference.
One-step and two-step prediction methods should be used according to whether a high accuracy of less than 1.5 mm or a low search and inference cost is desired.

\subsubsection{Sensitivity Analysis}
\label{sssec:sensitivity-analysis}
\begin{figure}
    \centering
    \includegraphics[width=0.95\linewidth]{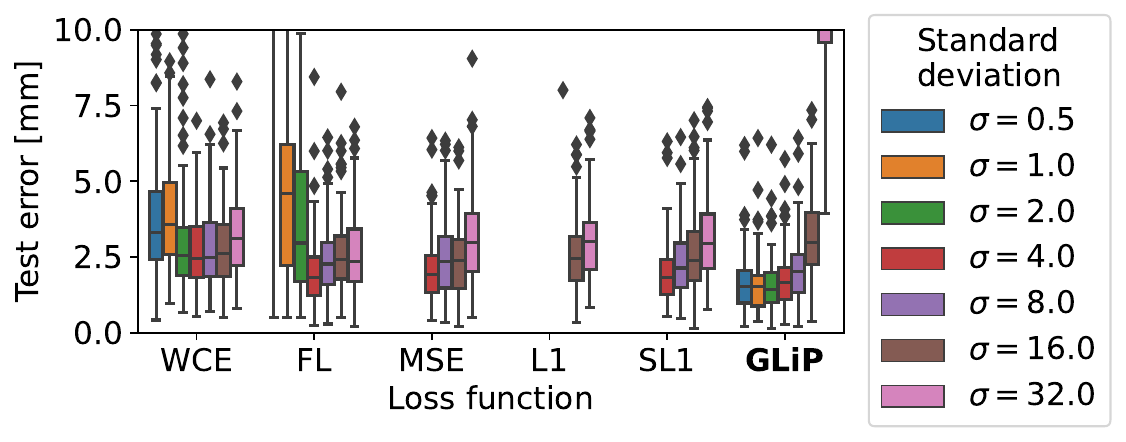}
    \caption{Boxplot of test Euclid distance error of predicted landmarks by UNet with loss functions for each standard deviation of input heatmap. Boxplots with test errors in the range greater than 10 are omitted.}
    \label{fig:heatmap-sigma}
\end{figure}
\begin{figure}
    \centering
    \includegraphics[width=0.9\linewidth]{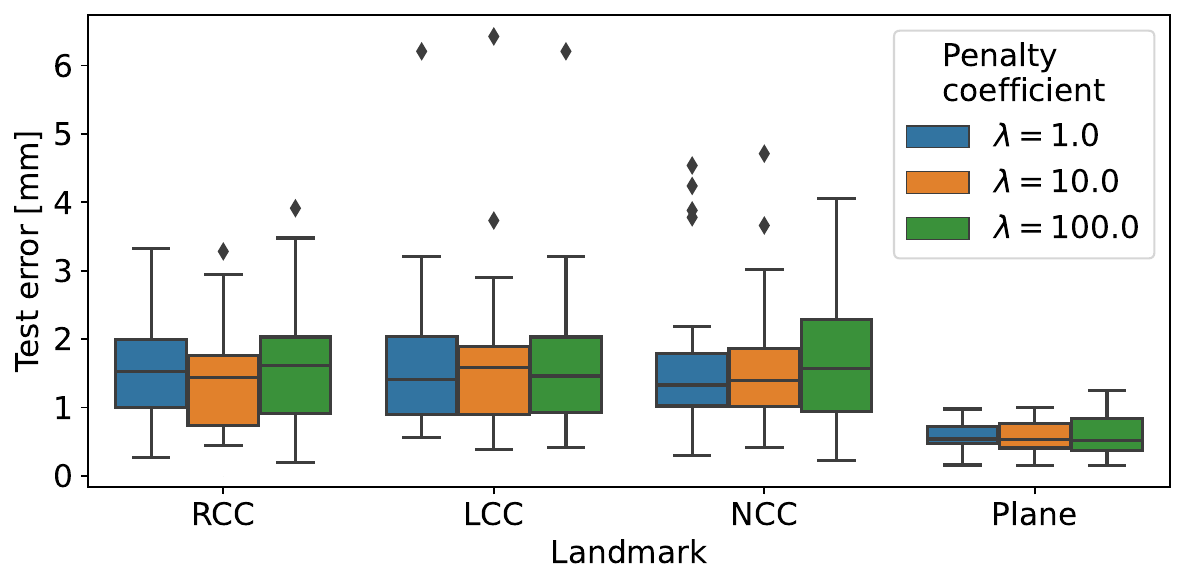}
    \caption{Boxplot of test Euclid distance error of predicted landmarks by UNet-GLiP for each penalty coefficient.}
    \label{fig:gradient-penalty}
\end{figure}

To verify the robustness of hyperparameters, a sensitivity analysis to the standard deviation of the heatmap $\sigma$ and the GLiP loss coefficient $\lambda$ was performed.
Because the main experiment's best CV median distance error is $\sigma=1.0$ and $\lambda=10$, we fixed $\lambda=10$ for the sensitivity analysis of its standard deviation and $\sigma=1.0$ for its coefficient.
Figure~\ref{fig:heatmap-sigma} shows the test Euclid error of the landmarks against their standard deviations.
GLiP's median error is below 2 for standard deviations between 0.5 and 4, while the prediction error rapidly worsens for standard deviations above 8.
Other loss functions record a minimum median error at standard deviations of 4 or 16.
Since the standard deviation corresponds to the uncertainty of the heatmap, it is difficult to predict more precisely than the standard deviation.
On the other hand, if the standard deviation is too small, it is difficult to learn properly due to data imbalance, which causes a large bias of data between binary values of 0 and 1 during training.
This is a trade-off between prediction precision in the standard deviation of the heatmap and learning stability.
For loss functions other than GLiP, the data imbalance destabilizes the learning process, resulting in large errors when the standard deviation is 1 or 2.
GLiP can decrease the landmark error with lower standard deviations because the optimal transport-type loss stabilizes the learning process.
Since even an expert can make an error of about 2 mm when annotating landmarks \cite{Astudillo20-segmentation}, setting the standard deviation to 1 or 2 is equivalent to constructing a heat map that only allows expert-level errors, and GLiP's standard deviation setting is aggressive.

Figure~\ref{fig:gradient-penalty} shows a boxplot of the test Euclid distance error of the landmarks and planes predicted by UNet-GLiP for the GLiP loss factor $\lambda$.
The median distance error is stable below 2 regardless of the coefficient.
The ablation study on the GLiP loss shows that the error is large when the coefficient is zero.
These results show that, regardless of the coefficients' values, the inclusion of a Lipschitz constraint term improves landmark prediction accuracy.

\section{Conclusion}
\label{sec:conclusion}

This paper proposes a new method for anatomical landmark prediction.
The proposed method incorporates a loss function with a Lipschitz constraint penalty corresponding to the voxel-wise classification problem based on optimal transport.
Through our experiments, the proposed method has a smaller Euclidean error of predicted landmarks than existing methods and other loss functions.
Surprisingly, the median error is below 2.0 mm, the expert's level, from images with a coarse resolution of 1.6 mm.
This is because the standard deviation of the heatmap minimizing CV error is lower than that of other loss functions, and the proposed method can achieve both accurate prediction and stable learning.

Despite this success, the paper has a few limitations.
The first limitation is that the dataset used is private.
We regret that we cannot give the reader the reproducibility of the experiments, and we are investigating the possibility of making some of the data sets publicly available.
The second limitation is that we have not applied the proposed method to architecture other than U-Net.
We plan to experiment with state-of-the-art architectures such as Swin-UNet \cite{cao22-swinunet}, ACC-UNet \cite{ibtehaz23-accunet}, and UNETR++ \cite{shaker24-unetr++}.

We plan to utilize the proposed method to estimate the prosthetic valve size, which is essential for TAVI planning.
Landmark localization of the aortic root gives three points through the aortic valve and the plane through these points.
If we can estimate the size of the patient's valve based on the three points on the estimated plane, we can determine the size of the prosthetic valve.
We will use image processing and machine learning to predict the prosthetic valve size from the estimated plane of the proposed method.

This paper can potentially have a wide range of impacts in the field of medical image analysis.
The proposed method can make accurate predictions from coarse CT images with a resolution of 1.6 mm and advances landmark localization for coarse images.
Due to the radiologist's skill, patient movement, or equipment deterioration, only coarse images can often be obtained.
Even when such bad data is only available, precise prediction can reduce the burden on physicians.
The experiments in this paper use only 90 to 91 data samples for training, contributing to landmark prediction for small datasets.
Many hospitals have only small datasets because they keep their datasets private, and publicly available datasets do not always match the format of their images.
Under such circumstances, the ability to perform highly accurate learning with a small amount of data (around 100 samples) is extremely significant.
Our proposed method can be applied to other devices, such as X-rays and MRIs, and other body parts, such as the skull and knees.
This will facilitate the development of the entire field of anatomical landmark localization.


\section*{Acknowledgement}
We are deeply grateful to Mr. Maruichi Jungo, Department of Radiology, Sendai Kousei Hospital, for collecting and managing CT data.
We would like to express my sincere gratitude to Mr. Shinichi Suzuki, Department of Radiology, Sendai Kousei Hospital, for helping us create ground-truth landmarks using his advanced CT analysis skills.
This work was partly supported by JSPS KAKENHI Grant Numbers JP21K20920, JP23K11156.

\bibliographystyle{ieeetr}
\bibliography{ref}

\end{document}